\documentclass{article}
\usepackage[utf8]{inputenc}
\usepackage[]{graphicx}
\usepackage[]{geometry}
\usepackage{color, soul}
\usepackage{iclr2017_workshop,times}
\usepackage{hyperref}
\usepackage{url}
\usepackage{caption}
\usepackage{subcaption}
\usepackage[fleqn]{amsmath}
\usepackage{amsfonts}
\newcommand{\ignore}[1]{}

\title{Discovering objects and their relations from entangled scene representations}

\author{\textbf{D. Raposo\thanks{Denotes equal contribution.}, \; A. Santoro\footnotemark[1], \; D.G.T. Barrett, \; R. Pascanu, \; T. Lillicrap, \; P. Battaglia}\\
DeepMind \\
London, United Kingdom \\
\small \texttt{\{draposo, adamsantoro, barrettdavid, razp, countzero, peterbattaglia\}@google.com}
}

\date{}

\newcommand{\comm}[1]{}
\newcommand{\e}[1]{{\mathbb E}}

\begin{document}
\maketitle

\begin{abstract}
Our world can be succinctly and compactly described as structured scenes of objects and relations. A typical room, for example, contains salient objects such as tables, chairs and books, and these objects typically relate to each other by their underlying causes and semantics. This gives rise to correlated features, such as position, function and shape. Humans exploit knowledge of objects and their relations for learning a wide spectrum of tasks, and more generally when learning the structure underlying observed data. In this work, we introduce \emph{relation networks} (RNs) - a general purpose neural network architecture for object-relation reasoning. We show that RNs are capable of learning object relations from scene description data. Furthermore, we show that RNs can act as a bottleneck that induces the factorization of objects from entangled scene description inputs, and from distributed deep representations of scene images provided by a variational autoencoder. The model can also be used in conjunction with differentiable memory mechanisms for implicit relation discovery in one-shot learning tasks. Our results suggest that relation networks are a potentially powerful architecture for solving a variety of problems that require object relation reasoning.
\end{abstract}

\section{Introduction}

The ability to reason about objects and relations is important for solving a wide variety of tasks \citep{spelke1992origins, lake2016building}. For example, object relations enable the transfer of learned knowledge across superficial dissimilarities \citep{tenenbaum2011grow}: the predator-prey relationship between a lion and a zebra is knowledge that is similarly useful when applied to a bear and a salmon, even though many features of these animals are very different. 

In this work, we introduce a neural network architecture for learning to reason about -- or model -- objects and their relations, which we call \emph{relation networks} (RNs). RNs adhere to several key design principles. First, RNs are designed to be invariant to permutations of object descriptions in their input. For example, RN representations of the object set $\{\mbox{table, chair, book} \}$ will be identical for arbitrary re-orderings of the elements of the set. Second, RNs learn relations \emph{across} multiple objects rather than \emph{within} a single object -- a basic defining property of object relations. This results from the use of shared computations across groups of objects. In designing the RN architecture, we took inspiration from the recently developed \textit{Interaction Network} (IN) \citep{Battaglia2016} which was applied to modelling physical, spatiotemporal interactions. Our model is also related to various other approaches that apply neural networks directly to graphs \citep{scarselli2009graph,bruna2013spectral,li2015gated,duvenaud2015convolutional,henaff2015deep,defferrard2016convolutional,kipf2016semi,edwards2016graph}.

In principle, a deep network with a sufficiently large number of parameters and a large enough training set should be capable of matching the performance of a RN. In practice, however, such networks would have to learn both the permutation invariance of objects and the relational structure of the objects in the execution of a desired computation. This quickly becomes unfeasible as the number of objects and relations increase.

Exploiting contextual relations among entities in scenes and other complex systems has been explored in various branches of computer science, biology, and other fields. In computer vision, \cite{zhu2009unsupervised} and \cite{zhao2011image} modelled relations among image features using stochastic grammars, \cite{felzenszwalb2008discriminatively} modelled relations among object parts using the ``deformable parts" model, and \cite{choi2012tree} modelled relations among objects in scenes using tree structured context models. In graphics, a number of approaches have been used to capture contextual scene structure, such as energy models \citep{yu2011make}, graphical and mixture models \citep{fisher2011characterizing,fisher2012example}, stochastic grammars \citep{liu2014creating}, and probabilistic programs \citep{talton2012learning,yeh2012synthesizing}. 

We used a scene classification task to evaluate RNs' ability to discover relations between objects. In this task, classification boundaries were defined by the relational structure of the objects in the scenes. There are various ways of encoding scenes as observable data; for example, scene description data can consist of sets of co-occurring objects and their respective features (location, size, color, shape, etc.). A typical room scene, then, might consist of the object set $\{\mbox{table, chair, lamp} \}$ and their respective descriptions (e.g., the table is large and red). We note that, although we allude to visual objects, such as tables and chairs, and speak of modeling visual relations, such as location and size, the datasets on which RNs can operate are not necessarily grounded in visual scenes. RNs are meant to model entities and relations, whether they are embedded in visual scenes, molecular networks, voting patterns, etc. %Similar datasets have been used for many decades in cognitive psychology to study human relation learning \citep{hemmer2009integrating} \adam{Peter add references}. 
Here we consider synthetic scene description data generated by hierarchical probabilistic generative models of objects and object features. 

We begin by motivating RNs as a general purpose architecture for reasoning about object relations. Critically, we describe how RNs implement a permutation invariant computation on implicit groups of factored ``objects." We then demonstrate the utility of RNs for classification of static scenes, where classification boundaries are defined by the relations between objects in the scenes. Next, we exploit RNs' implicit use of factored object representations to demonstrate that RNs can induce the factorization of objects from entangled scene inputs. Finally, we combine RNs with memory-augmented neural networks (MANNs) \citep{santoro2016one} to solve a difficult one-shot learning task, demonstrating the ability of RNs' to act in conjunction with other neural network architectures to rapidly discover new object relations from entirely new scenes.

\section{Model}

\subsection{Description and implementation}
RNs are inspired by \textit{Interaction Networks} (INs) \citep{Battaglia2016}, and therefore share similar functional insights. Both operate under the assumption that permutation invariance is a necessary requirement for solving relational reasoning problems in a data efficient manner. However, INs use relations between objects as input to determine object \textit{interactions}, mainly for the purpose of reasoning about dynamics. RNs compute object relations, and hence aim to determine object-relational structure from static inputs.

Suppose we have an object $o_i = (o_i^1, o_i^2, ..., o_i^n)$, represented as a vector of $n$ features encoding properties such as the object's type, color, size, position, etc. A collection of $m$ objects can be gathered into a $m \times n $ matrix $D$, called the \textit{scene description}. Although the term \textit{scene description} alludes to visual information, this need not be the case; scenes can be entirely abstract, as can the objects that constitute the scene, and the features that define the objects. 

We can imagine tasks (see section \ref{sec:experimental_tasks}) that depend on the \textit{relations}, $r$, between objects. One such task is the discovery of the relations themselves. For example, returning to the predator and prey analogy, the predator-prey relation can be determined from \textit{relative} features between two animals -- such as their relative sizes, perhaps. Observation of the size of a single animal, then, does not inform whether this animal is a predator or prey to any other given animal, since its size necessarily needs to be compared to the sizes of other animals.

\begin{figure}[ht]
	\centering
	\includegraphics[width=0.8\textwidth]{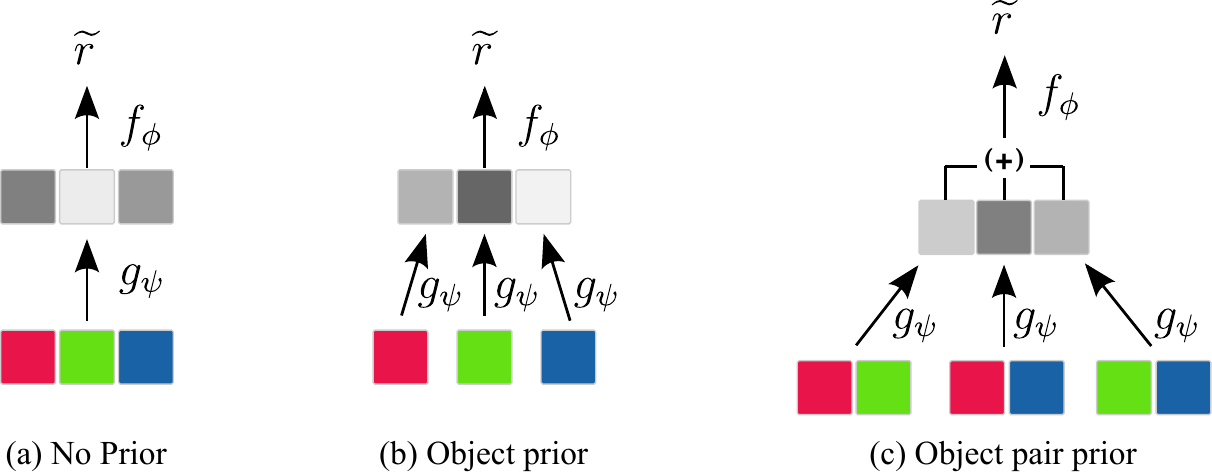}
    \caption{\textbf{Model types.} RNs are constructed to operate with an explicit prior on the input space (c). Features from all pairwise combinations of objects act as input to the same MLP, $g_{\psi}$.}
    \label{fig:model_description}
\end{figure}

There are a number of possible functions that could support the discovery of object relations (figure \ref{fig:model_description}). Consider a function $g$, with parameters $\psi$. The function $g_{\psi}$ can be defined to operate on a particular factorization of $D$; for example, $g_{\psi}(D) \equiv g_{\psi}(o_{1}^2, ...,o_i^j, ...,o_m^n)$. We are interested in models generally defined by the composite form $ f \circ g $, where $f$ is a function that returns a prediction $\widetilde{r}$. 

One implementation of $g_{\psi}$ would process the entire contents of $D$ without exploiting knowledge that the features in a particular row are related through their description of a common object, and would instead have to learn the appropriate parsing of inputs and any necessary sub-functions: $\widetilde{r} = f_{\phi}(g_{\psi}(o_1^1, o_1^2, ..., o_m^n))$. An alternative approach would be to impose a prior on the parsing of the input space, such that $g$ operates on objects directly: $\widetilde{r} = f_{\phi}(g_{\psi}(o_1), g_{\psi}(o_2), ..., g_{\psi}(o_m))$. A third middle-ground approach -- which is the approach taken by RNs -- recognizes that relations necessarily exist in the context of a set of objects that have some capacity to be related. Thus, the computation of relations should entail some common function across these sets. For example, $g$ may operate on \textit{pairs} of objects: $\widetilde{r} = f_{\phi}(g_{\psi}(o_1, o_2), g_{\psi}(o_1, o_3), ..., g_{\psi}(o_{m-1}, o_m))$. For RNs, $g$ is implemented as a multi-layered perceptron (MLP) that operates on pairs of objects. The same MLP operates on all possible pairings of objects from $D$.

The second defining architectural property of RN's is object permutation invariance. To incorporate this property, we further constrain our architecture with an aggregation function $a$ that is commutative and associative: $\widetilde{r} = f_{\phi}(a(g_{\psi}(o_1, o_2), g_{\psi}(o_1, o_3), ..., g_{\psi}(o_{m-1}, o_m)))$. The order invariance of the aggregation function is a critical feature of our model, since without this invariance, the model would have to learn to operate on all possible permuted pairs of objects without explicit knowledge of the permutation invariance structure in the data. A natural choice for $a$ is summation. Thus, the model that we explore is given by $\widetilde{r} =  f_{\phi}(\sum_{i, j} g_{\psi}(o_i, o_j))$ where $f_{\phi}$ and $g_{\psi}$ are MLP's that we optimise during training.

\section{Experimental tasks and data}
\label{sec:experimental_tasks}

\subsection{Datasets}
\begin{figure}[h]
	\centering
	\includegraphics[width=0.95\textwidth]{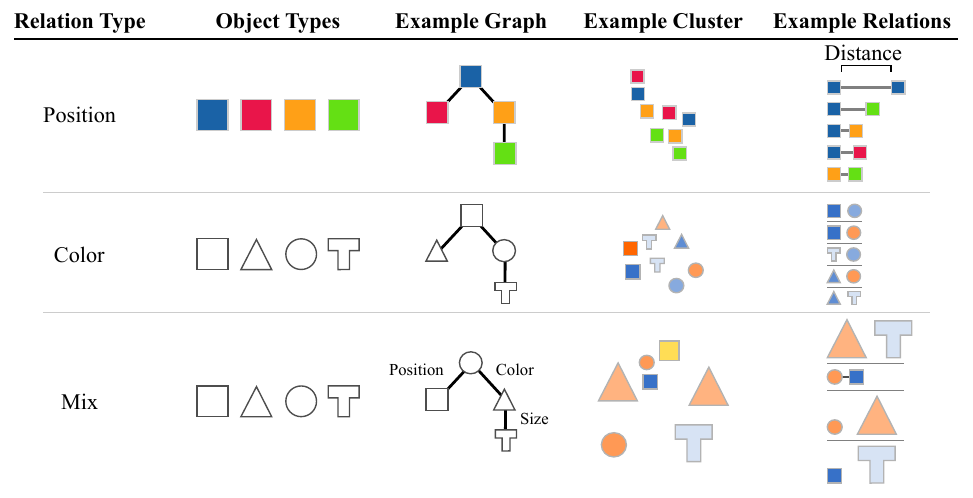}
    \caption{\textbf{Objects and relations.} Relation types (column one) between object types (column two) can be described with directed graphs (column three). Shown in the fourth column are cropped clusters from example scenes generated by a model based on the directed graph shown in column three. In the last column are example of relations that can be used to inform class membership; for example, the distances between pairs of objects, or the differences in color between pairs of objects that may inform the particular graphical structure, and hence generative model, used to generate the scene.}
    \label{fig:object_example}
\end{figure}

To probe a model's ability to both infer relations from scene descriptions and implicitly use relations to solve more difficult tasks -- such as one-shot learning -- we first developed datasets of scene descriptions and their associated images. To generate a scene description, we first defined a graph of object relations (see figure \ref{fig:object_example}). For example, suppose there are four types of squares, with each type being identified by its color. A graph description of the relations between each colored square could identify the blue square as being a parent of the orange square. If the type of relation is ``position," then this particular relation would manifest as blue squares being independently positioned in the scene, and orange squares being positioned in close proximity to blue squares. Similarly, suppose we have triangles and circles as the object types, with color as the relation. If triangles are parents to circles, then the color of triangles in the scene will be randomly sampled, while the color of a circle will be derived from the color of the parent triangle. Datasets generated from graphs, then, impose solutions that depend on relative object features. That is, no information can be used from \textit{within} object features alone -- such as a particular coordinate position, or RGB color value.

Graphs define generative models that can be used to produce scenes. For scenes with position relations, root node coordinates $(o_x^p, o_y^p)$, were first randomly chosen in a bounded space. Children were then randomly assigned to a particular parent object, and their position was determined as a function of the parent's position:  $(o_x^c, o_y^c) =  (o_x^p + d \cos(\theta^c), o_y^p + d \sin(\theta^c))$. Here, $\theta^c \sim \mathcal{U}(\theta^p - \pi/3, \theta^p +\pi/3)$, where $\theta^p$ is the angle computed for the parent. For root nodes, $\theta^p \sim \mathcal{U}(0, 2\pi)$. $d$ is a computed distance: $d = d_0 + d_1$, where $d_0$ is a minimum distance to prevent significant object overlap, and $d_1$ is sampled from a half-normal distribution. This underlying generative structure -- inherit features from parents and apply noise -- is used to generate scenes from graphs that define other relations, such as color. For the case of color, the inherited features are RGB values. Ultimately, scene descriptions consisted of matrices with 16 rows, with each row describing the object type: position, color, size and shape (four rows for each of four types).

Custom datasets were required to both explicitly test solutions to the task of inferring object relations, and to actively control for solutions that do not depend on object relations. For example, consider the common scenario of child objects positioned close to a parent object, analogous to chairs positioned around a table. In this scenario, the count information of objects (i.e., that there are more child objects than parent objects) is non-relational information that can nonetheless be used to constrain the solution space; the prediction of the relation between objects doesn't entirely depend on explicitly computed relations, such as the relative distance between the child objects and the parent objects. To return to the kitchen scene analogy, one wouldn't need to know the relative distance of the chairs to the table; instead, knowing the number of chairs and number of tables could inform the relation, if it is known that less frequently occurring objects tend to be parents to more frequently occurring objects. Although information like object count can be important for solving many tasks, here we explicitly sought to test models' abilities to compute and operate on object relations.

The datasets will be made freely available. 

\subsection{Task descriptions}
\label{sec:task_descriptions}
The tasks with which we assessed the RN's performance fell into three categories. The first category involved the classification of scenes. In this task, the network was given a scene description as input, with the target being a binary matrix description of the edges between object types (this is known as the graph adjacency matrix) -- see figure \ref{fig:object_example}. Training data consisted of 5000 samples derived from 5, 10, or 20 unique classes (i.e., graphs), with testing data comprising of withheld within-class samples. Although the target was a matrix description of the generating graph, this task was fundamentally one of classification. Nonetheless, since class membership can only be determined from the relational structure of the objects within a particular sample, the ability to classify in this way is dependent on the ability to infer relations.  

\begin{figure}[h]
	\centering
	\includegraphics[width=0.8\textwidth]{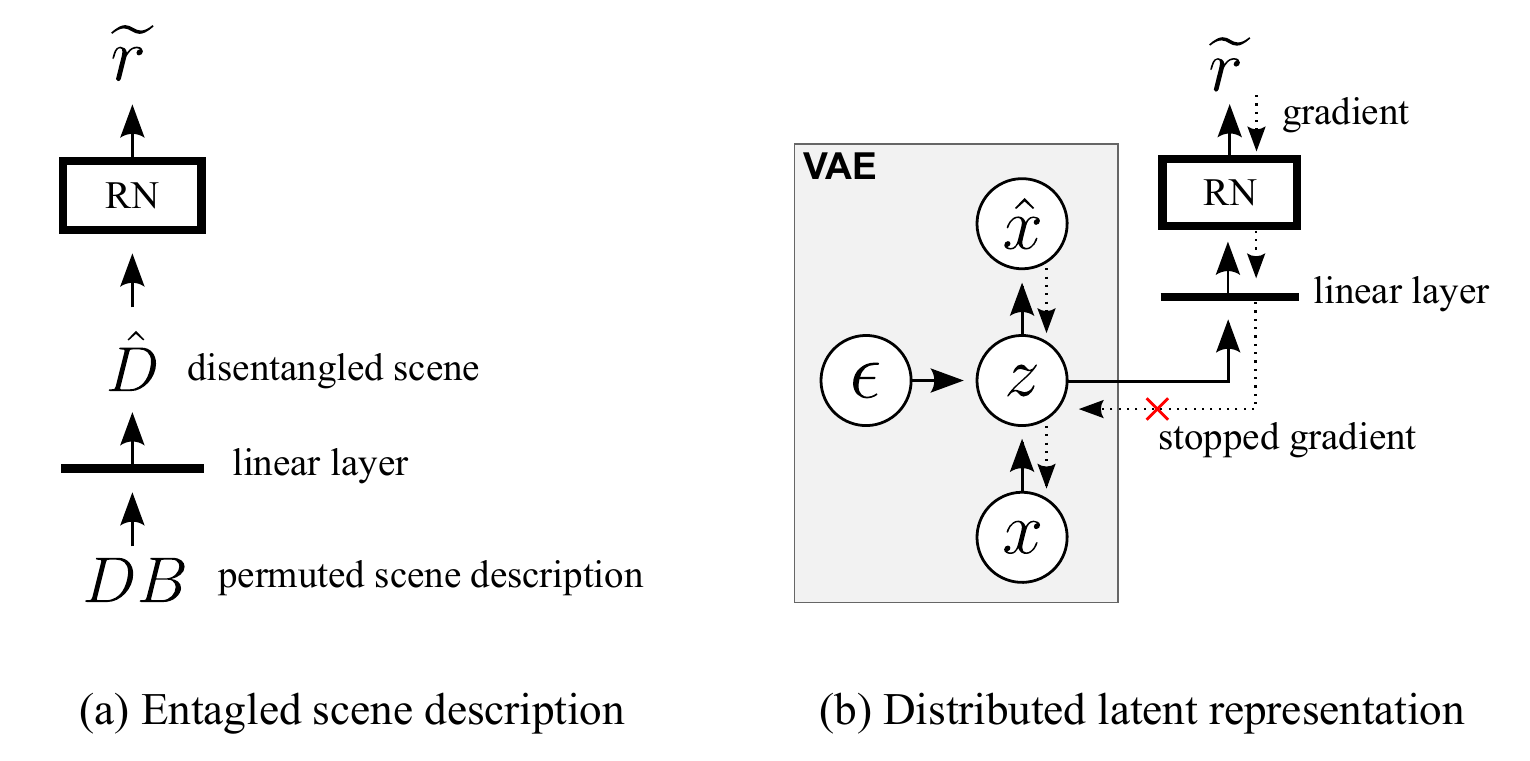}
    \caption{\textbf{Scene entangling}. To test the ability of the RN to operate on entangled scene representations, we (a) multiplied a flattened vector representation of the scene description by a fixed permutation matrix $B$, or, (b) passed pixel-level representations of the scenes through a VAE and used the latent code as input to a RN with an additional linear layer.}
    \label{fig:shuffle_models}
\end{figure}

The second category of tasks tested the ability of the RN to classify scenes from unstructured input domains (see figure \ref{fig:shuffle_models}). Since RNs operate on factored object representations, this task specifically probed the ability of RNs to \textit{induce} the learning of object factorizations from entangled scene descriptions. In the first set of experiments we broke the highly structured scene description input by passing it through a fixed permutation matrix. The RN must now learn to reason about a randomly permuted object feature representation. To decode the entangled state we used a linear layer whose output provided the input to the RN. We then asked the network to classify scenes, as in the previous tasks. In the second set of experiments we pushed this idea further, and tested the ability of the RN to classify scenes from pixel-level representations. To enable this, images were passed through a variational autoencoder (VAE), and the latent variables were provided as input to a RN.

The final category of tasks tested the implicit use of discovered relations to solve a difficult overarching problem: one-shot relation learning \citep{vinyals2016matching, santoro2016one, lake2015human}. In one-shot relation learning, a memory augmented RN must learn to classify a novel scene in one-shot (after a single presentation of a novel scene from our data-set). To train a memory augmented RN for one-shot relation learning, sequences of samples were fed to a memory-augmented neural network (MANN) with a relational network pre-processor. Sequences -- or \textit{episodes} -- consisted of 50 random samples generated from five unique graphs, from a pool of 1900 total classes, presented jointly with time-offset label identifiers, as per \citet{hochreiter2001learning} and \citet{santoro2016one}. Critically, the labels associated with particular classes change from episode-to-episode. Therefore, the network must depend on within-episode knowledge to perform the task; it must learn the particular label assigned to a class within an episode, and learn to assign this label to future samples of the same class, within the same episode. Labels are presented as input in a time-offset manner (that is, the correct label for the sample presented at time $t$ is given as input to the network at time $t+1$) to enable learning of an arbitrary binding procedure. Unique information from a sample -- which is necessarily something pertaining to the relations of the contained objects -- must be extracted, bound to its associated label, and stored in memory. Upon subsequent presentations of samples from this same class, the network must query its memory, and use stored information to infer class membership. There is a critical difference in this task compared to the first. In this task, identifying class labels change constantly from episode-to-episode. So, the network cannot simply encode mappings from certain learned relational structures to class labels. Instead, the only way the network can solve the task is to develop an ability to compare and contrast extracted relational structures between samples as they occur within an episode. Please see the appendix for more details on the task setup.

\section{Additional model components and training details}

\subsection{Variational autoencoder}
\label{sec:vae}
For inferring relations from latent representations of pixels we used a variational autoencoder (VAE) \citep{kingma2013auto} with a convolutional neural network (CNN) as the feature encoder and deconvolutional network as the decoder (see figure \ref{fig:shuffle_models}b). The CNN consisted of two processing blocks. The input to each block was sent through four dimension preserving parallel convolution streams using 8 kernels of size 1x1, 3x3, 5x5, and 7x7, respectively. The outputs from these convolutions were passed through a batch normalization layer, rectified linear layer and  concatenated. This was then convolved again with a down-sampling kernel of size 3x3, halving the dimension size of the input, and, except for the final output layer, again passed through batch normalization and rectified linear layers. The entire CNN consisted of two of these blocks positioned serially. Therefore, input images of size 32x32 were convolved to feature-maps of size 8x8. The feature decoder consisted of these same blocks, except convolution operations were replaced with deconvolution operations.

The final feature maps provided by the CNN feature encoder were then passed to a linear layer, whose outputs constituted the observed variables $x$ for the VAE. $x$, which was decomposed into $\mu$ and $\sigma$, was then used with an auxiliary Gaussian noise variable $\epsilon$ to infer the latent variables $z = \mu + \epsilon \sigma$. These latent variables were then decoded to generate the reconstruction $\hat{x}$, as per the conventional implementation of VAEs. Along a separate pathway, the latent variable representation $z$ of an image was fed as input to a linear layer, which projected it to a higher dimensional space -- the scene description $D$ (see figure \ref{fig:shuffle_models}b). Importantly, this connection -- from $z$ to $D$ -- did not permit the backward flow of gradients to the VAE. This prevented the VAE architecture from contributing to the RN's disentangling solution. 

\subsection{Memory-augmented neural network}
\label{sec:MANN}
For implicit discovery of relations from scene descriptions, the RN was used as a pre-processor for a memory-augmented neural network (MANN). The MANN was implemented as in \citet{santoro2016one}, and the reader is directed here for full details on using networks augmented with external memories. Briefly, the core MANN module consists of a controller -- a long-short term memory (LSTM) \citep{hochreiter1997long} -- that interacts with read and write heads, which in turn interact with an external memory store. The external memory store consists of a number of memory slots, each of which contains a vector ``memory." During reading, the LSTM takes in an input and produces a query vector, which the read head uses to query the external memory store using a cosine distance across the vectors stored in the memory slots, and returns a weighted sum of these vectors based on the cosine distance. During writing, the LSTM outputs a vector that the write head uses to write into the memory store using a least recently used memory access mechanism \citep{santoro2016one}.

\subsection{Training details}
The sizes of the RN -- in terms of number of layers and number of units for both $f_{\phi}$ and $g_{\psi}$ -- were \{200, 200\}, \{500, 500\}, \{1000, 1000\}, or \{200, 200, 200\}. We also trained a MLP baseline (for the sake of comparison) using equivalent network sizes. We experimented with different sizes for the output of $g_{\psi}$. Performance is generally robust to the choice of size, with similar results emerging for 100, 200, or 500. The MANN used a LSTM controller size of 200, 128 memory slots, 40 for the memory size, and 4 read and write heads. 

The Adam optimizer was used for optimization \citep{kingma2014adam}, with learning rate of $1e^{-4}$ for the scene description tasks, and a learning rate of $1e^{-5}$ for the one-shot learning task. The number of iterations varied for each experiment, and are indicated in the relevant figures. All figures show performance on a withheld test-set, constituting 2-5\% of the size of the training set. The number of training samples was 5000 per class for scene description tasks, and 200 per class (for 100 classes) for the pixel disentangling experiment. We used minibatch training, with batch-sizes of 100 for the scene description experiments, and 16 (with sequences of length 50) for the one-shot learning task.

\section{Results}

\subsection{Supervised learning of relational structure from scene descriptions}
\label{sec:supervised_sd}

\begin{figure}[ht]
	\centering
    \includegraphics[width=\textwidth]{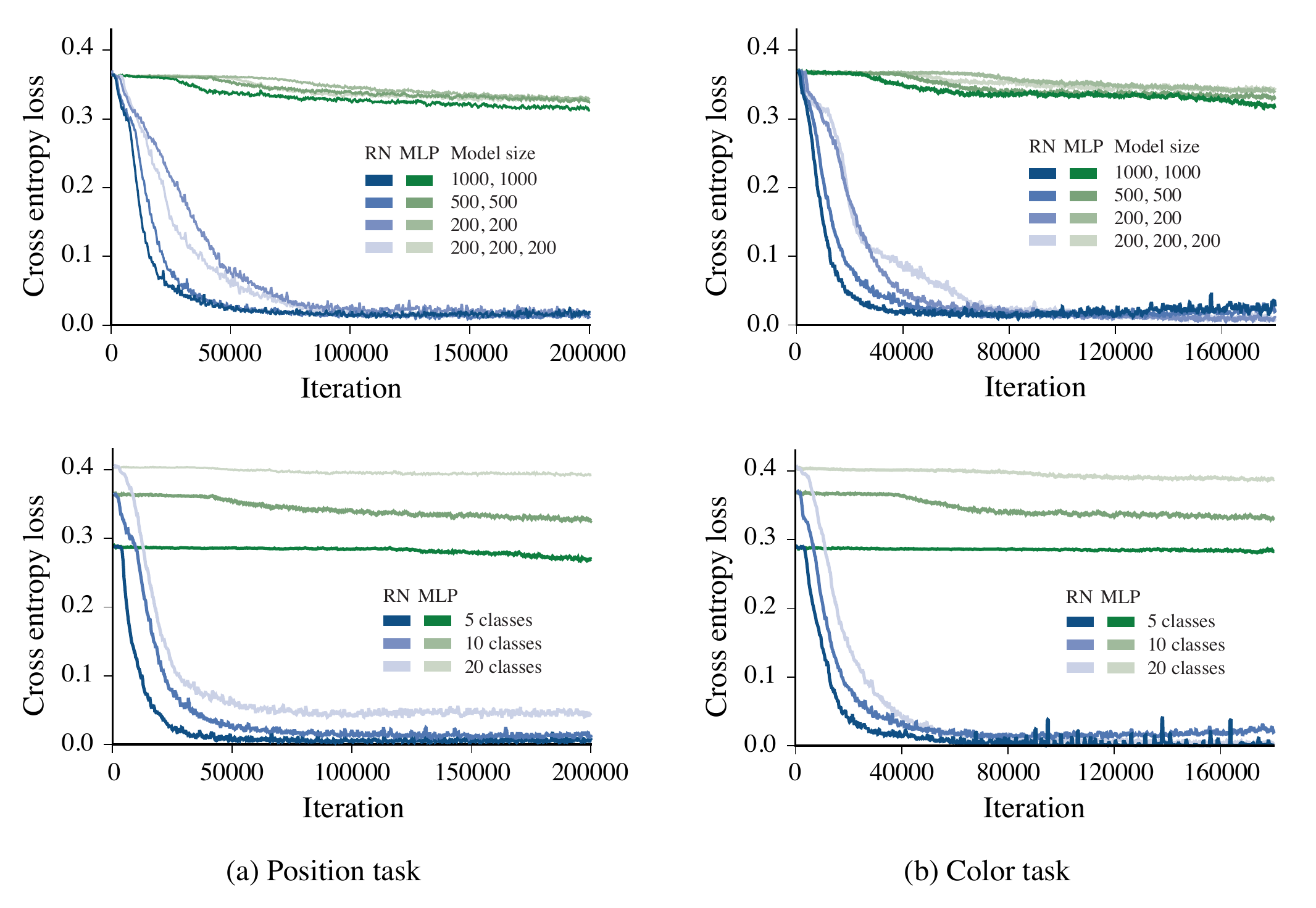}
    \caption{\textbf{Scene classification tasks.} (a) RNs of various sizes (legend, inset) performed well when trained to classify 10 scene classes based on position relations, reaching a cross entropy loss below 0.01 (top panel), and on tasks that contained 5, 10 or 20 classes (bottom panel). The MLPs performed poorly regardless of network size and the number of classes. (b) When relational structure depended on the color of the objects (\textit{color task}), all RN configurations performed well classifying 5, 10 or 20 classes, similar to what we observed on the position task. MLPs with similar number of parameters performed poorly.}
    \label{fig:position_color_tasks}
\end{figure}

We begin by exploring the ability of RNs to learn the relation structure of scenes, by training RNs to classify object relations (the adjacecy matrix) of a scene description, as descriped in section  \ref{sec:task_descriptions}. As a baseline comparison, we contrasted their performance with that of MLPs of different sizes and depths. First, we compared the performance of these models on scenes where the relational structure was defined by position (figure \ref{fig:position_color_tasks}a). After 200,000 iterations the RNs reached a cross entropy loss of ~0.01 on a withheld test set, with MLPs of similar size failing to match this performance (figure \ref{fig:position_color_tasks}a, top). In fact, the smallest RN performed significantly better than the largest MLP. The performance of the MLPs remained poor even after 1 million iterations (cross entropy loss above 0.2 -- not shown). This result was consistent for datasets with 5, 10 and 20 scene classes (figure \ref{fig:position_color_tasks}a, bottom).

\begin{figure}[ht]
	\centering
	\includegraphics[width=0.5\textwidth]{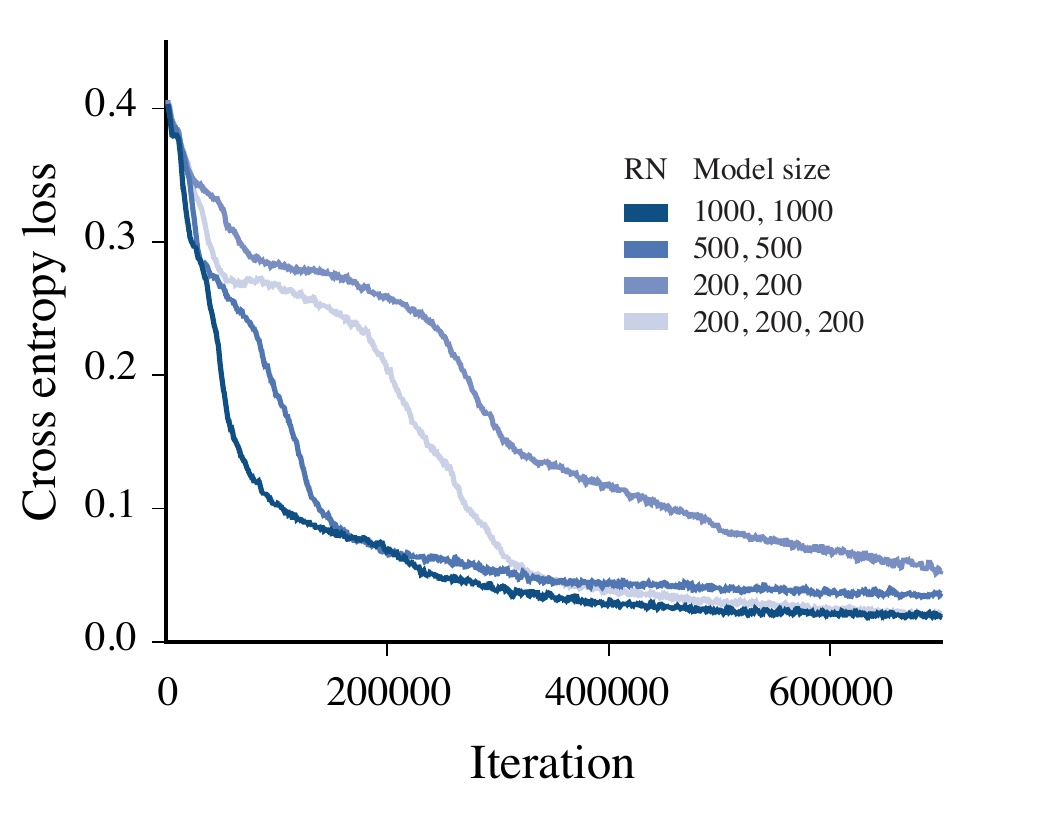}
    \caption{\textbf{Scene classification on withheld classes.} RNs of different sizes (legend, inset) were trained to classify scenes from a pool of 490 classes. The plot shows the cross entropy loss on a test set composed of samples from 10 previously unseen classes.}
    \label{fig:unseen_classes}
\end{figure}

These results were obtained using targets that explicitly described the relations between objects (the adjacency matrix). Using one-hot vectors as output targets for classification resulted in very similar results, with the RNs' accuracy reaching 97\% (see figure \ref{fig:onehot_class} in appendix). We use adjacency matrix graph descriptions as a target for learning because this target explicitly probes that ability of the RN model to represent relations between individual pairs of object types independent of other relations that might be present in a scene. Explicitly targeting the individual relations could in principle allow the model to learn the particular components that form the overall scene structure. Indeed, when trained to classify hundreds of scene classes using the adjacency matrix for specifying the class, the RN was then able to generalize to unobserved classes (see figure \ref{fig:unseen_classes}). (Note: this generalization to unseen classes would be impossible with the use of one-hot labels for classification targets, since there is no training information for unseen classes in that case). Additionally, the ability to generalize to unobserved classes suggests that the RN is able to generalize in a combinatorially complex object-relation space because of its ability to learn compositional object-relation structure. It is able to use pieces (i.e., specific relations) of learned information and combine them in unique, never-before-seen ways, which is a hallmark feature of compositional learning.

We repeated this experiment using relational structure defined by the color of the objects, with position being randomly sampled for each object (figure \ref{fig:position_color_tasks}b). Again, RNs reached a low classification error on withheld data (below 0.01), whereas the MLP models did not (error above 0.2; figure \ref{fig:position_color_tasks}b, top). This was observed across datasets with 5, 10 and 20 classes (figure \ref{fig:position_color_tasks}b, bottom).

\subsection{Inferring object relations from non-scene description inputs}
Next, we explore the ability of RNs to infer object relations directly from input data that is not nicely organized into a scene description matrix of factored object representations. This is a difficult problem because RNs require a scene description matrix of objects as input. In most applications, a scene description matrix is not directly available, and so, we must augment our RN with a mechanism for transforming entangled representations (such as pixel images) directly into a representation that has the properties of a scene description matrix. Specifically, we require a transformation that produces an $m\times n$ dimensional matrix $D$ whose rows can be interpreted as objects, and whose columns can be interpreted as features.

In this section, we explore architectures that can support object relation reasoning from entangled representations. We also explore the extent to which RNs can act as an architectural bottleneck for inducing object-like representations in the RN input layer. We will consider two types of datasets: pixel image representations and entangled scene descriptions.

\subsubsection{Inferring relations from entangled scene descriptions}
In this task we probed the network's ability to classify scenes from entangled scene descriptions. Intuitively, the RN should behave as an architectural bottleneck that can aid the disentangling of objects by a downstream perceptual model to produce factored object representations on which it can operate. To create an entangled scene description data-set we started from the scene description matrix $D$, and reshaped it into a vector of size $mn$ (i.e., a concatenated vector of objects $[o_1; o_2; o_3; ..; o_m]$). We then transformed this vector with a random permutation matrix $B$ of size $mn \times mn$. We chose a permutation matrix for entanglement because it preserves all of the input information without scaling between factors, and also, because it is invertable. Therefore, it should be possible to train a linear layer to disentangle the entangled scenes into a format suitable for RN's.

Following this, we augment our RN with a downstream linear layer positioned before the RN and after the entangled scene data. The linear layer is represented by a learnable matrix $U$ of size $mn \times mn$ (without biases) (see figure \ref{fig:shuffle_models} (a)). This linear layer produces a $mn$ dimensional vector as output. This output vector is  reshaped into a matrix of size $m \times n$ before being provided as input to the RN. 

\begin{figure}[h]
	\centering
    \includegraphics[width=\textwidth]{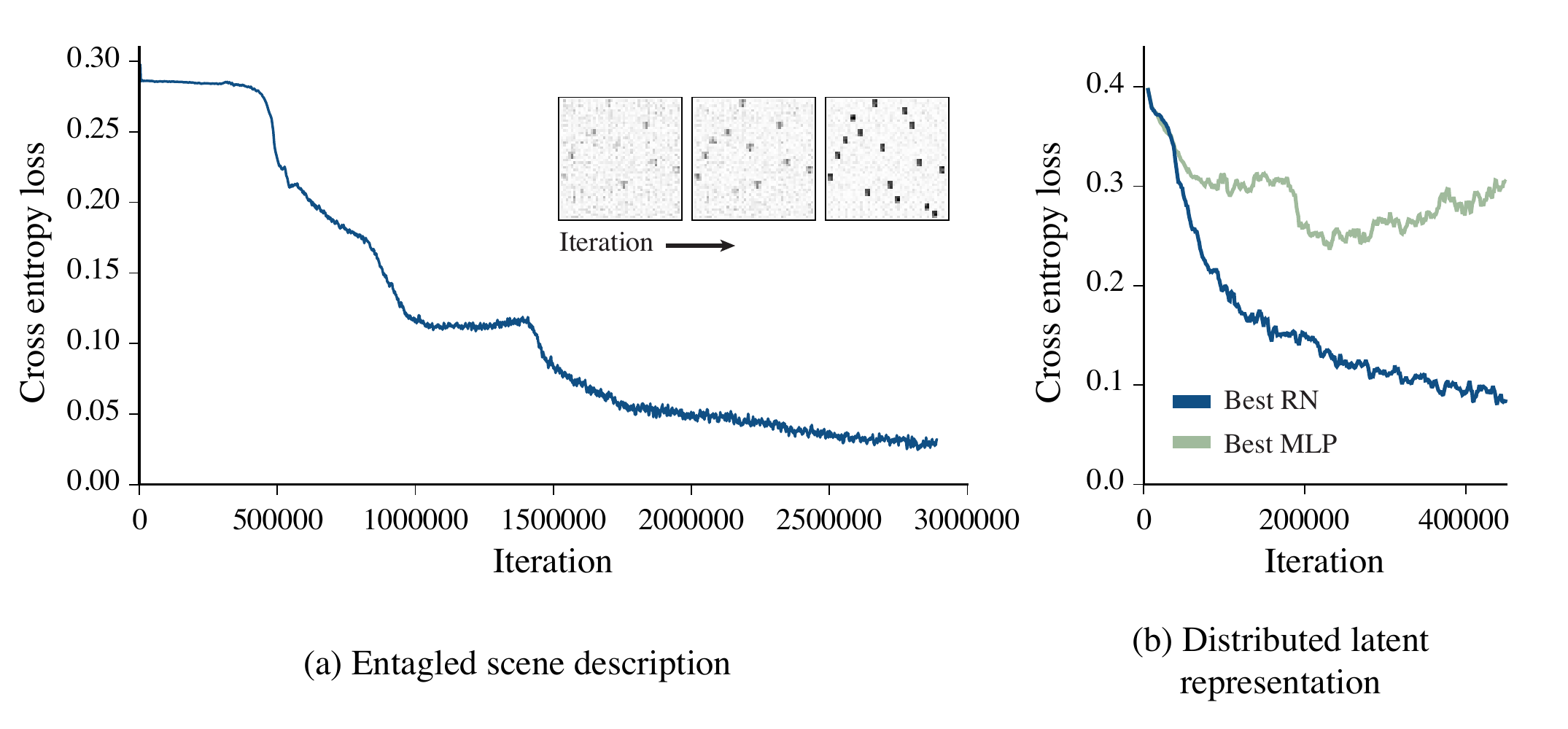}
    \caption{\textbf{Inferring relational structure from entangled scenes.} (a) A RN augmented with a linear layer $U$ is able to reason about object relations from a scene description matrix that has been entangled with a random permutation matrix $B$. Examples of matrix $UB$ (inset) during training (for a single example seed) illustrates that the linear layer learns to recover factorised object representations. (b) Distributed latent representations are inferred latent variables from VAE processing of image depictions of the scenes.}
    \label{fig:deshuffle}
\end{figure}

This augmented RN successfully learns to classify entangled scene descriptions (figure \ref{fig:deshuffle}a). During training, the linear layer learns to provide a factorised representation of $m$ objects to the RN (figure \ref{fig:deshuffle}a, inset). To illustrate this we visualized the absolute value of $UB$ (figure \ref{fig:deshuffle}a, inset). We noticed a block structure emerging (where white pixels denote a value of 0, and black pixels denote a value of 1). The block structure of $UB$ suggests that object $k$, as perceived by the network in the linear output layer, was a linear transformation (given by the block) of object $l$ from the ground truth input in $D$. This gradual discovery of new objects over time is particularly interesting. It is also interesting to note that since the RN is order invariant, there is no pressure to recover the ground truth order of the objects. We can see that this is the case, because the matrix $UB$ is not block-diagonal. Also, the exact order of features that define the object were not disentangled, since there is no pressure from the RN architecture for this to happen. In this way, RNs manage to successfully force the identification of different objects without imposing a particular object ordering or feature ordering \textit{within} objects.

\subsubsection{Inferring relations from pixels}
Next, we explore the ability of the RN to classify scenes from pixel image representations of scenes. We use image depictions from 100 unique scene classes from our position-relation dataset (figure \ref{fig:example_scenes} in appendix).

Image representations of scenes cannot be directly fed to a RN because the objects in images are not represented in a factorised form. We need to augment our RN with an image preprocessing mechanism that has the capacity to produce factorised object representations. For these purposes, we augment our RN with a variational autoencoder (VAE), whose latent variables provide input to a linear layer, which in turn provides input to the RN (similar to the linear layer in the previous section) (figure \ref{fig:shuffle_models} (b)). Both the VAE and RN were trained concurrently -- however, gradients from the RN were not propagated to the VAE portion of the model. The VAE learns to produce a entangled latent representation of the input images. There is no pressure for the VAE to produce a representation that has factorised objects. Instead, as in the previous section, the linear layer must learn to disentagle the image representation into a object-factorised representation suitable for the RN.

By preventing gradients from the RN to propogate through to the VAE portion of the model, we prevent any of the VAE components from contributing to the solution of the RN's task. Thus, this experiment explicitly tested the ability of the RN to operate on distributed representations of scenes. It should be noted that this task is different from the previous entangling tasks, as the VAE, in principle, may be capable of providing both disentangled object representations, as well as relational information in its compressed latent code. Nonetheless, the results from this task demonstrate the capacity for RNs to operate on distributed representations, which opens the door to the possible conjunction of RNs with perceptual neural network modules (figure \ref{fig:deshuffle}b). 

\subsection{Implicit use of relations for one-shot learning}
Finally, we assess the potential to use RNs in conjunction with memory-augmented neural networks to quickly -- and implicitly -- discover object relations and use that knowledge for one-shot learning.

We trained a MANN with a RN pre-processor to do one-shot classification of scenes, as described in section \ref{sec:task_descriptions}. In order to solve this task, the network must store representations of the scenes (which, if class representations are to be unique, should necessarily contain relational information), and the episode-unique label associated with the scene. Once a new sample of the same class is observed, it must use inferred relational information from this sample to query its memory and retrieve the appropriate label.

\begin{figure}[h]
	\centering
	\includegraphics[width=0.94\textwidth]{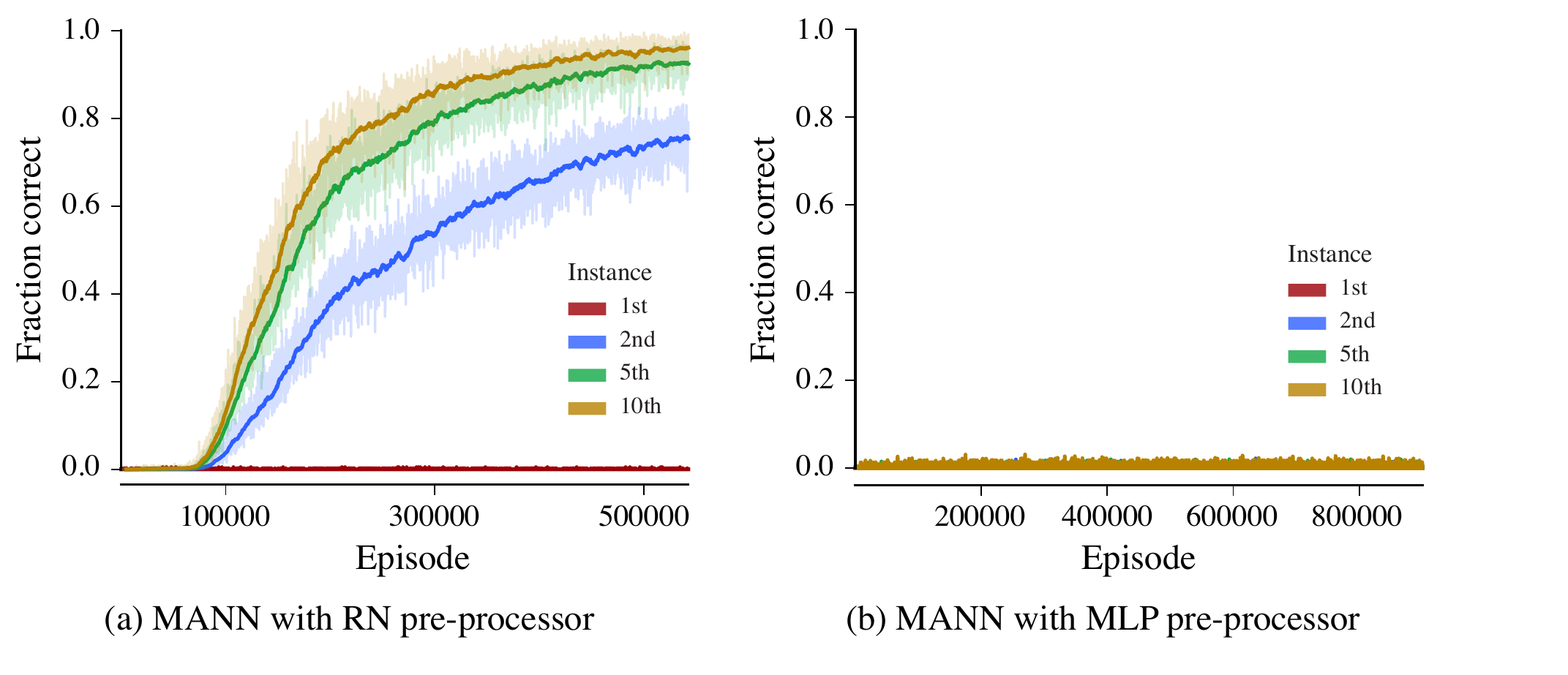}
    \caption{\textbf{One-shot classification of scenes.} (a) A MANN with a RN pre-processor is able to accurately classify unobserved scenes after the presentation of a single instance of the same class. (b) A MANN with a MLP pre-processor performs at chance.}
    \label{fig:meta_learning}
\end{figure}

During training, a sequence of 50 random samples from 5 random classes (out of a pool of 1900) were picked to constitute an episode. The test phase consisted of episodes with scenes from 5 randomly selected and previously unobserved classes (out of a pool of 100). After 500,000 episodes the network showed high classification accuracy when presented with just the second instance of a class (76\%) and performance reached 93\% and 96\% by the 5th and 10th instance, respectively (figure \ref{fig:meta_learning}a). As expected, since class labels change from episode-to-episode, performance is at chance for the first instance of a particular class.

Although samples from a class are visually very dissimilar, a RN coupled to an external memory was able to do one-shot classification. This suggests that the network has the capacity to quickly extract information about the relational structure of objects, which is what defines class membership, and does so without explicit supervision about the object-relation structure. Replacing the RN pre-processor with an MLP resulted in chance performance across all instances (figure \ref{fig:meta_learning}b), suggesting that the RN is a critical component of this memory-augmented model.

\section{Conclusions}

RNs are a powerful architecture for reasoning about object-relations. Architecturally, they operate on pairs of objects (or object-like entities) and they enforce permutation invariance of objects. This architecture naturally supports classification of object-relation structure in scenes. RNs can also be augmented with additional modules to support a wider range of tasks. For example, we have demonstrated that RNs can be augmented with MANNs to support one-shot relation learning. 

RNs can be used in conjunction with perceptual modules for reasoning about object-relations in images and other entangled representations. In doing so, they can induce factored object representations of entangled scenes. The form of these factored ``objects" can be quite flexible, and may permit representations of inputs that are not localised spatially, or that consist of multiple disjoint entities. In future work, it would be interesting to explore the range of ``objects" that RN's augmented with perceptual modules can discover across a variety of datasets.

The utility of the RN as a relation-reasoning module suggests that it has the potential to be useful for solving tasks that require reasoning not only about object-object relations, but also about verb-object relations, as in human-object interaction datasets \citep{chao2015hico} or question-answering tasks that involve reasoning between multiple objects \citep{krishna2016visual}. Leveraging RNs for these problem domains represents an exciting direction for future investigation.

\subsubsection*{Acknowledgments}
We would like to thank Scott Reed, Daan Wierstra, Nando de Freitas, James Kirkpatrick, and many others on the DeepMind team.

\bibliographystyle{iclr2017_conference}
\bibliography{iclr2017_conference}

\section*{Appendix}

\subsection*{Additional scene classification experiments}

\begin{figure}[ht]
	\centering
	\includegraphics[width=0.5\textwidth]{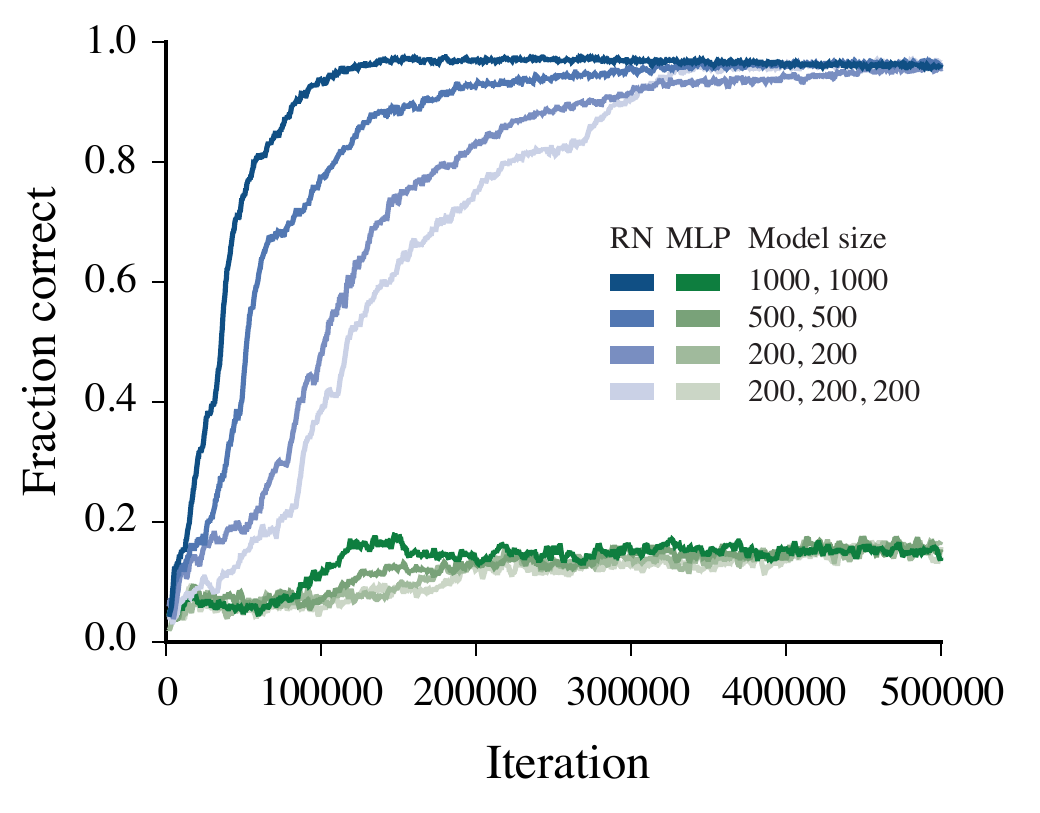}
    \caption{\textbf{Scene classification with one-hot vector labels.} Performance of RNs and MLPs on the \textit{position task} with 20 classes and using one-hot vectors as output targets.}
    \label{fig:onehot_class}
\end{figure}

\subsection*{One-Shot Learning Task Setup}
The one-shot learning task proceeds as in \cite{hochreiter2001learning} and \cite{santoro2016one}. In this task, parameters $\theta$ are updated to reduce the expected learning cost across a distribution of datasets $p(\mathcal{D})$:
\begin{align*}
    \theta^* = \text{arg }\text{min}_{\theta} \e{E}_{\mathcal{D} \sim p(\mathcal{D})} [\mathcal{L(\mathcal{D};\theta)}]
\end{align*}

\begin{figure}[ht]
	\centering
	\includegraphics[width=0.6\textwidth]{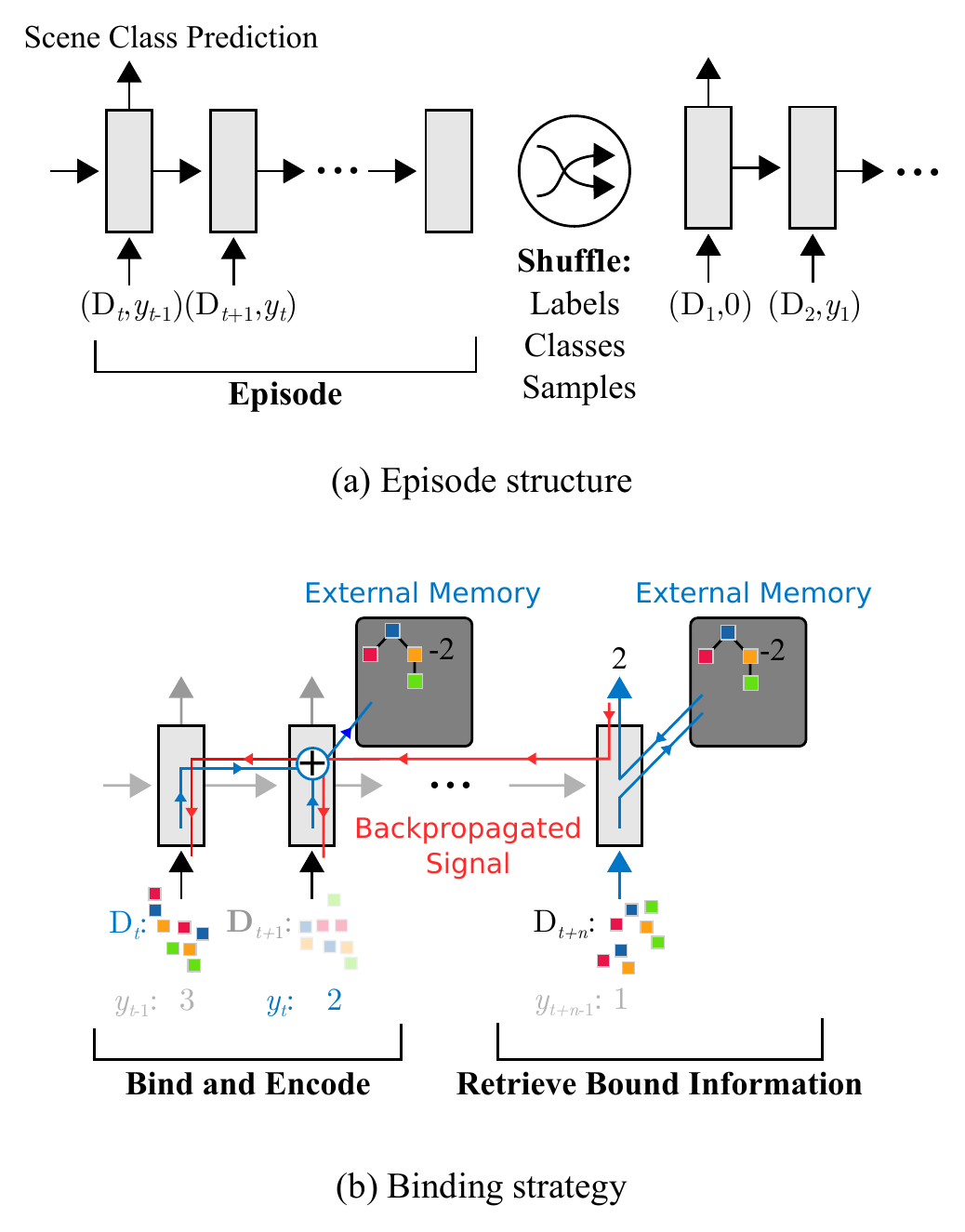}
    \caption{\textbf{One-shot learning task setup}}
    \label{fig:one-shot-task}
\end{figure}

In our case, datasets consist of sequences, or episodes, of input-output pairs, where inputs are scene descriptions $D$ and outputs are target labels $y$ (see figure \ref{fig:one-shot-task}). So, $\{(D_t, y_t)\}_{t=1}^T \in \mathcal{D}$. All $D$ from all datasets are constructed using the same generative process for producing scene descriptions, as described in the main text. The main differentiating factor between datasets, then, is the labels associated with each scene description. From dataset-to-dataset, scenes from a particular scene-class are assigned a unique, but arbitrary target label. This implies that scenes from the same scene class will not necessarily have the same label from episode-to-episode (or equivalently, from dataset-to-dataset), but will indeed have the same label \emph{within} episodes. 

\begin{figure}[ht]
	\centering
	\includegraphics[width=0.7\textwidth]{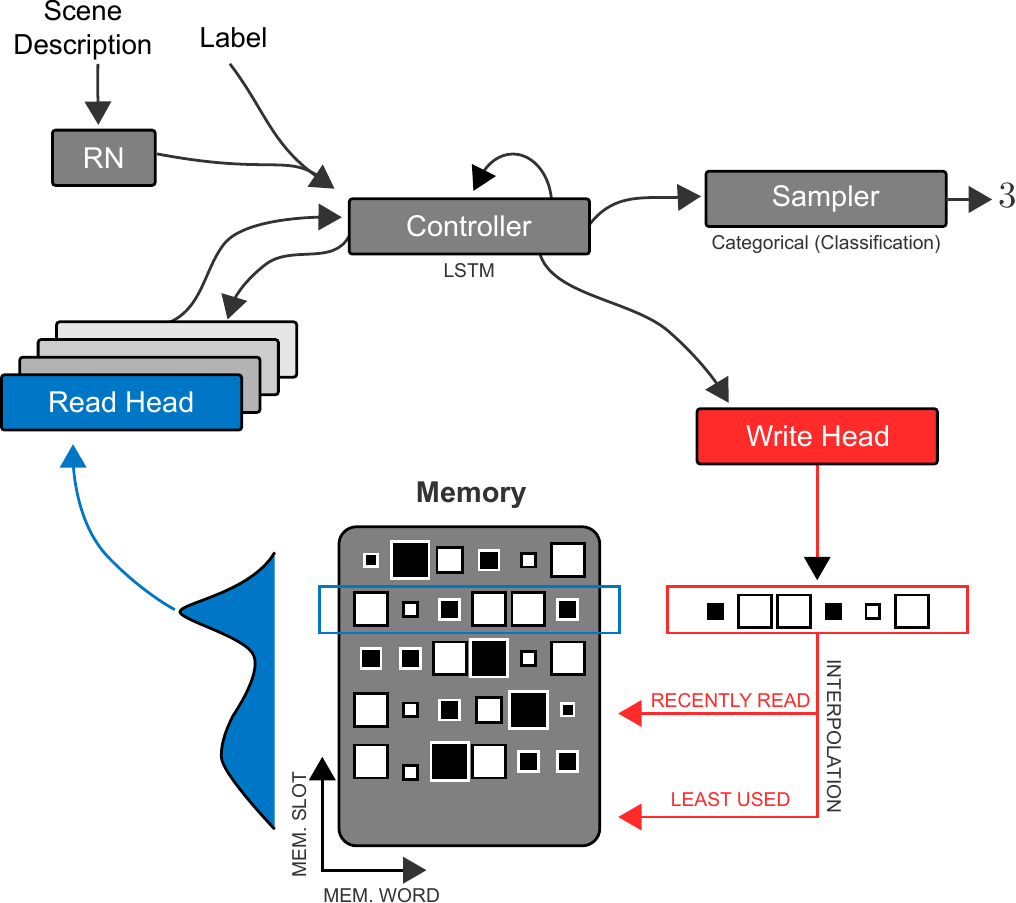}
    \caption{\textbf{RN with MANN}}
    \label{fig:mann}
\end{figure}

In this training setup, labels are provided as input at subsequent timesteps. So, a given input at a particular timestep is $(D_t, y_{t-1})$, where $y_{t-1}$ is the correct target label for the previous timestep's scene description. This allows the MANN to learn a binding strategy \citep{santoro2016one} (see figure \ref{fig:one-shot-task} and \ref{fig:mann}): it will produce a representation for a particular input and bind it with the incoming label, and will then store this bound information in memory for later use. Importantly, the label is presented in a time offset manner; if labels were instead presented at the same timestep as the corresponding sample, then the network could learn to cheat, and use the provided label information to inform its output.

This particular setup has been shown to allow for a particular meta-learning strategy \citep{hochreiter2001learning,santoro2016one}; the network must learn to bind useful scene representations with arbitrary, episode-specific labels, and use this bound information to infer class membership for subsequent samples in the episode. This task poses a particular challenge for RNs; they must have the capacity to quickly extract useful representations of scene classes to enable rapid, within-episode comparisons to other scene representations. The representation extracted for a single example must contain useful, general information about the class if it is to provide information useful for subsequent classification. As seen in the results, one-shot accuracy is quite high, implying that the RN and MANN combination can extract useful representations of scene classes given a single example, and use this stored representation to successfully classify future examples from the same class. This implies that the network is necessarily extracting relational information from input scenes, since only this information can allow for successful one-shot learning across samples. 

\begin{figure}[ht]
	\centering
	\includegraphics[width=0.7\textwidth]{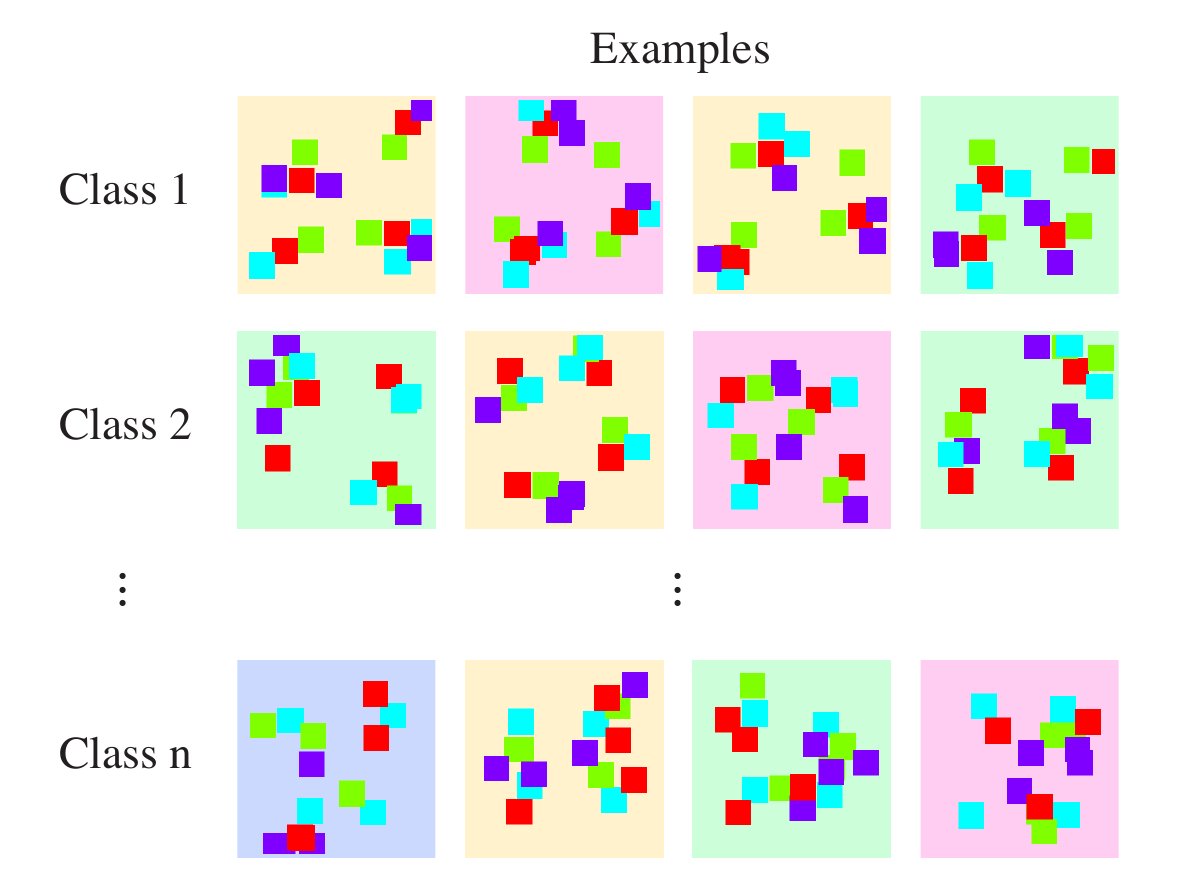}
    \caption{\textbf{Example scene depictions}}
    \label{fig:example_scenes}
\end{figure}

\end{document}